\documentclass[10pt,twocolumn,letterpaper]{article}

\usepackage{wacv}
\usepackage{times}
\usepackage{epsfig}
\usepackage{graphicx}
\usepackage{amsmath}
\usepackage{amssymb}

\wacvfinalcopy 

\ifwacvfinal
\def\assignedStartPage{1} 
\fi

\ifwacvfinal
\usepackage[breaklinks=true,bookmarks=false]{hyperref}
\else
\usepackage[pagebackref=true,breaklinks=true,colorlinks,bookmarks=false]{hyperref}
\fi

\ifwacvfinal
\else
\fi

\begin{document}

\title{Deep Learning for Fitness}

\author{Mahendran N\\
Indian Institute of Technology Tirupati\\
Andhra Pradesh, India\\
{\tt\small cs19m008@iittp.ac.in}}

\maketitle

\begin{abstract}

We present Fitness tutor, an application for maintaining correct posture during workout exercises or doing yoga. Current work on fitness focuses on suggesting food supplements, accessing workouts, workout wearables does a great job in improving the fitness. Meanwhile, the current situation is making difficult to monitor workouts by trainee. Inspired by healthcare innovations like robotic surgery, we design a novel application Fitness tutor which can guide the workouts using pose estimation. Pose estimation can be deployed on the reference image for gathering data and guide the user with the data. This allow Fitness tutor to guide the workouts (both exercise and yoga) in remote conditions with a single reference posture as image. We use posenet model in tensorflow with p5js for developing skeleton. Fitness tutor is an application of pose estimation model in bringing a realtime teaching experience in fitness. Our experiments shows that it can leverage potential of pose estimation models by providing guidance in realtime.
   
\end{abstract}

\section{Introduction}


During these unprecedented times we are bound to be in home without much interaction to rest of the world. Staying physically active in stay-at-home period is a challenging task \cite{Nyenhuis20}. Practicing workout exercises or yoga can improve overall body health. The results will be exponential with proper guidance or with a mentor. However, this current situation makes it impossible for guiding physical work via video call as entire posture can't be seen in 2D frame.

Recent inclusion of technology in fitness, can increase the workout results in less time. Technology can be used in fitness like workout applications, wearables , video conferencing with mentors. They analyse from the data collected from applications\cite{Leijdekkers15}, fitness machines\cite{Kulkarni19} and improvising the methods can produce better results. All of the existing applications play as support to the physical workouts, improvising physical workouts after workout is done. This is the major limitation of existing applications based on fitness. Doing exercise in home is good but without a coach, the posture, breathe in-out time may not be right. There are multiple implications like body pain, muscle soreness due to extra stress on muscles, wrong posture and so on. There are multiple methods that eliminates this problems even in mobile application like making live coaching, explaining the best posture in detail. Live coaching is possible, time flexibility can also be achieved. In live coaching individual care has to be taken while exercising or yoga but it can be costly and lot of time is needed for individual care. We propose an aspect of providing guidance to physical workout in real-time. The video conferencing guidance can be done for many physical workouts. This is highly effective approach. But it is not suitable for every exercise. Some exercises that exist in social media can be done without training but the correct posture should be maintained while doing the exercise. It is impossible to mentor every time when you train. Real-time suggestions are needed to improvise then and there and to get maximum output.

We propose Fitness tutor, an application to guide you when doing workouts. This applies pose estimation using deep learning to learn the body posture while exercising. Using video chats for posture correction with the mentor on other end, applications that provide animated real person exercising in video followed by voice commands may help but it wont predict the posture is right or wrong. We propose the method of correcting the posture with and without coach. We utilize deep learning PoseNet model in order to help correct the posture. The coach training image can be compared with the current posture that yoga enthusiast is doing. If the hand positions are wrong, the posenet model could identify with comparison with coach's posture and tells the user where they needs to correct. With the help of AI posenet model, this is possible when exercising.

The proposed approach can be considered as application of PoseNet model in tensorflow in best way possible. PoseNet model accurately identifies the poses of human body parts. With the help of pose estimation model we extends the usage to fitness application. With our proposed approach, user can perform the exercise and this online tutor helps in adjusting the posture.

The main contributions of this proposed paper are
\begin{itemize}
    \item Posture correction in real-time without human intervention with reference image.
    \item Reference images can be of their own trainer, online workouts.
    \item We propose this as an web application using ml5js posenet model. Simple and no installation is needed.
\end{itemize}


\section{Related Work}

\subsection{Fitness applications}
\label{fitnessapp}
Exercising can have direct impact on body and it is constant. Doing exercise not only enhances the physical aspect but also the mental aspect of human health. Doing workouts provide great relief to the mental stress. It increases productivity and feels fresh in every day to day activities. It increases creativity \cite{Steinberg97,Ramocki02,Roman18}, mental health, life expectancy \cite{Lee12} and decreases health problems \cite{Rominger20}. Steinberg et al.\cite{Steinberg97} research tells us that exercise enhances the positive mood and decreased the negative mood. Existing fitness applications can help in weight loss \cite{Pellegrini15,Pagoto13}, nutrition and diet, relaxation and general work out information \cite{Higgins16}. Researches suggest that participation in elite sport is generally favorable to lifespan longevity \cite{Lemez15}. Doing workouts regularly can decrease almost all of the physical problems and gives an upperhand for dealing with the world mentally and physically. 

\subsection{Deep learning for Health}

Researches in healthcare field using AI are booming. Reserach taking various forms like identification of disease in prior, classification into subcategories, analysing with pre existing medical data, combining healthcare with robotics using deep learning paves the way to save human lives. Deep learning also implemented in fitness applications like fitbit, smart watches and mobile applications to monitor activity. This data can help predicting specialized suggestions for achieving their goals. We take the idea of implementing deep learning in fitness for impacting lives directly.

\subsection{Action recognition}
Understanding of the human actions and recognising the particular behaviour is a complex task. Existing researches in this field emerges with $98.69\%$ accuracy models \cite{Kalfaoglu20} for action recognition dataset UCF101 \cite{Soomro12}. Action recognition plays an important role in human computer interaction, robotics. In our proposed method, since the pose estimation has to be in real time and quicker response, we use the ml5js posenet model for action recognition.

\section{Proposed Approach}
\subsection{Pose estimation}
Pose estimation in humans using Deep learning research gives upperhand in designing and developing applications based on human pose. Pose estimation is useful in tracing the body parts of human. This helps in knowing the alignment of human body parts and knowing human characteristics better from images. This opens up an opportunity in developing applications based on human pose comparison,
body part elevation and alignment.

The main idea of recognising the human actions with the body pose estimation and tracking can open up to a real world applications for scene understanding and vision applications like fitness. This approach completely depends on the pose estimation model \cite{Papandreou17,Papandreou18}. Their effort in proposing a pose estimation model and Dan Oved making it available for tensorflow.js makes this application for fitness possible. The tensorflow posenet model identifies 17 body parts as skeleton. Among them we use only certain body parts for comparison. Table \ref{tab:fittut} shows the body parts used for comparing the reference image with user data.

\begin{table}
    \centering
    \begin{tabular}{|c|c|c|}\hline
        S.No & PoseNet model & Fitness tutor \\
        1 & nose &  \\[0.25em]
        2 & leftEye &  \\[0.25em]
        3 & rightEye  & \\[0.25em]
        4 & leftEar  & \\[0.25em]
        5 & rightEar  & \\[0.25em]
        6 & leftShoulder & \checkmark \\[0.25em]
        7 & rightShoulder & \checkmark  \\[0.25em]
        8 & leftElbow &  \\[0.25em]
        9 & rightElbow &  \\[0.25em]
        10 & leftWrist & \checkmark \\[0.25em]
        12 & rightWrist &  \checkmark \\[0.25em]
        13 & leftHip & \checkmark \\[0.25em]
        13 & rightHip & \checkmark \\[0.25em]
        14 & leftKnee &  \\[0.25em]
        15 & rightKnee &  \\[0.25em]
        16 & leftAnkle & \checkmark \\[0.25em]
        17 & rightAnkle & \checkmark \\[0.25em] \hline
    \end{tabular}
    \caption{Table represents the body joints that are present in tensorflow and the joints that are used for comparison by fitness tutor}
    \label{tab:fittut}
\end{table}

\begin{figure*}
    \begin{center}
    \includegraphics[width=0.9\linewidth, height=10cm]{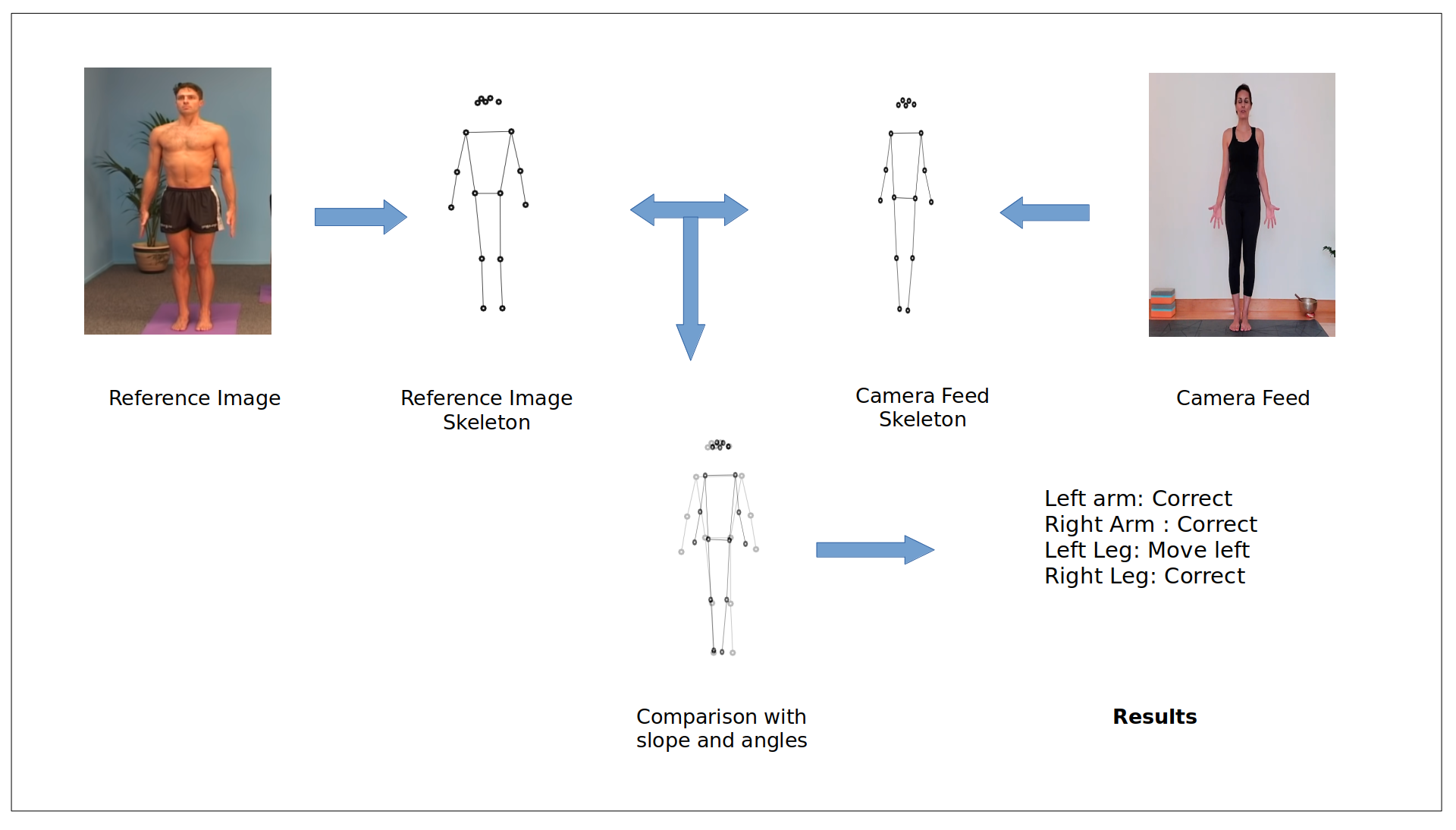}
    \caption{This represents the workflow of the Fitness Tutor application. Get the referenced image and extract the skeleton. It also gets the camera feed frame and extract skeleton from input. Both the skeleton are compared with specific body joints using slope made between them. Results are made according to the slope for the body parts.}
    \label{fig:architecture}
    \end{center}
\end{figure*}

We use the posenet model developed in ml5js library. p5js helps in utilizing posenet model and doing calculations on the results. The scene understanding and pose estimation are fetched using their approach. Our proposed approach consists of pose estimation for referenced image and for user input.
The exercising posture made by professionals is compared with the current posture done by the user. We gather the professional exercising posture and developed the skeletons in prior. When the user tries the pre defined posture by professional, camera fetches
the live user data and develop the skeleton for user. We use mathematical approach to compare the images. The resultant skeleton for reference image is stored as JSON file. With the referenced skeleton we match it with the user skeleton generated in real-time for suggesting the corrections. The architecture diagram of the Fitness tutor is shown in figure \ref{fig:architecture}. Comparison between the body skeleton is done on the basis of slope between coordinates.

\subsection{Dataset}

This method of doing exercise with professional and comparing the posture with them does not have an existing dataset. We depend on online platform in order to fetch the posture images for comparison. Scraping the social media with keywords like exercise, workout results in images which are mostly not related, whole body is not covered in order to get the entire posture. So we utilize properly documented youtube videos and extract postures from randomnly chosen youtube videos for comparison. Again in this social media we cant authenticate the exercise posture done by professionals or not. We assume some of the posture are authentic. Then we compare the other posture in order to meet the goal of finding the wrong one. This approach is similar to live camera feed. The live feed from p5js is converted to frames and only each frame is checked then and there. The results are provided for each frame. 

\subsection{Slope in body parts}

Slope between the coordinates is calculated between some of the body parts. Some of the slopes are not needed. For example, the slope between left and right eye is unnecessary and there is not much effect even if it is measured. Thus we find slopes of some parts which have enough data about the user. Slope between two coordinates are calculated using this formula,

\begin{center}
    $ Slope = \frac{(y2-y1)}{(x2-x1)}$
\end{center}

With the data, we find slope using above formula to match the posture upto $\pm0.5$ value. Table \ref{tab:majorpositions} tells the slope between coordinates are calculated to match with the referenced skeleton.

\begin{table}
    \centering
    \begin{tabular}{|c|c|c|}\hline
        Bodypart A & Bodypart B \\
        \hline
        Left shoulder & Left elbow \\
        Right shoulder & Right elbow  \\
        Left hip & Left ankle \\
        Right hip & Right ankle \\
        \hline
    \end{tabular}
    \caption{Table represents the body parts slope to be noted for pose comparison and tutoring}
    \label{tab:majorpositions}
\end{table}

From the tracking of 17 body parts by pose estimation model, we use specific body parts to match the pose. Slope between posture can help in training the user to match the exact posture. Since the pose estimation models provide high accurate position, the user cannot replicate the reference posture. So the slope values are accepted with a difference of $\pm 0.5$ which do not have drastic impact in posture. We provide suggestions based on slope. For example, if one arm is moved away from referenced posture, user is notified to move the arm up/down according to the slope value. 

\begin{figure*}[hbt!]
    
    \includegraphics[width=0.14\linewidth,height=4cm]{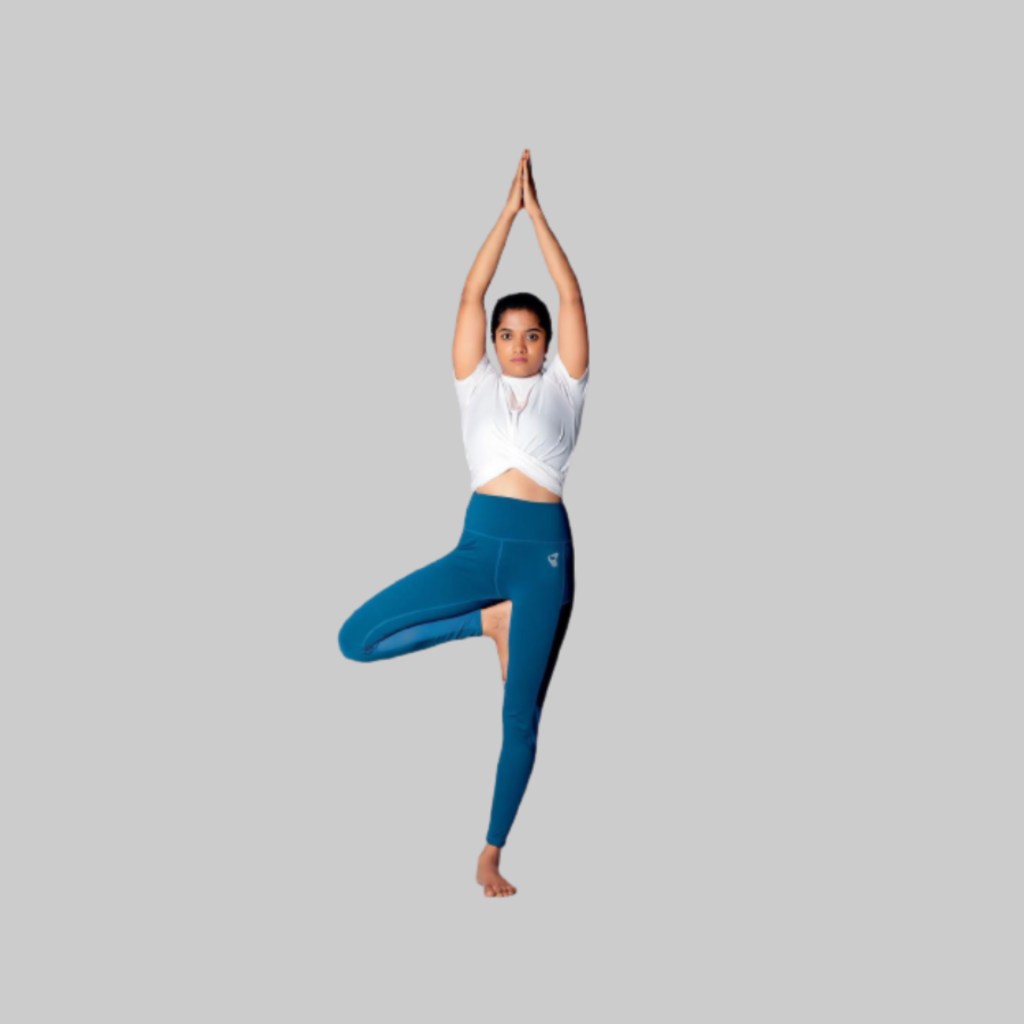}
    \hspace{0.25cm}
    \includegraphics[width=0.13\linewidth,height=4cm]{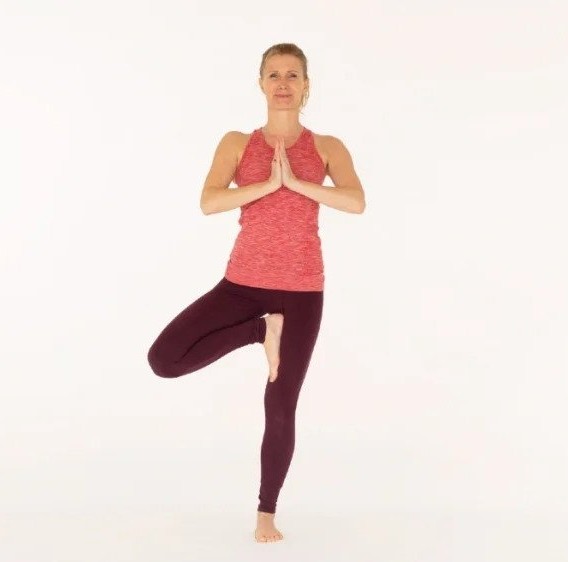}
    \includegraphics[width=0.13\linewidth,height=4cm]{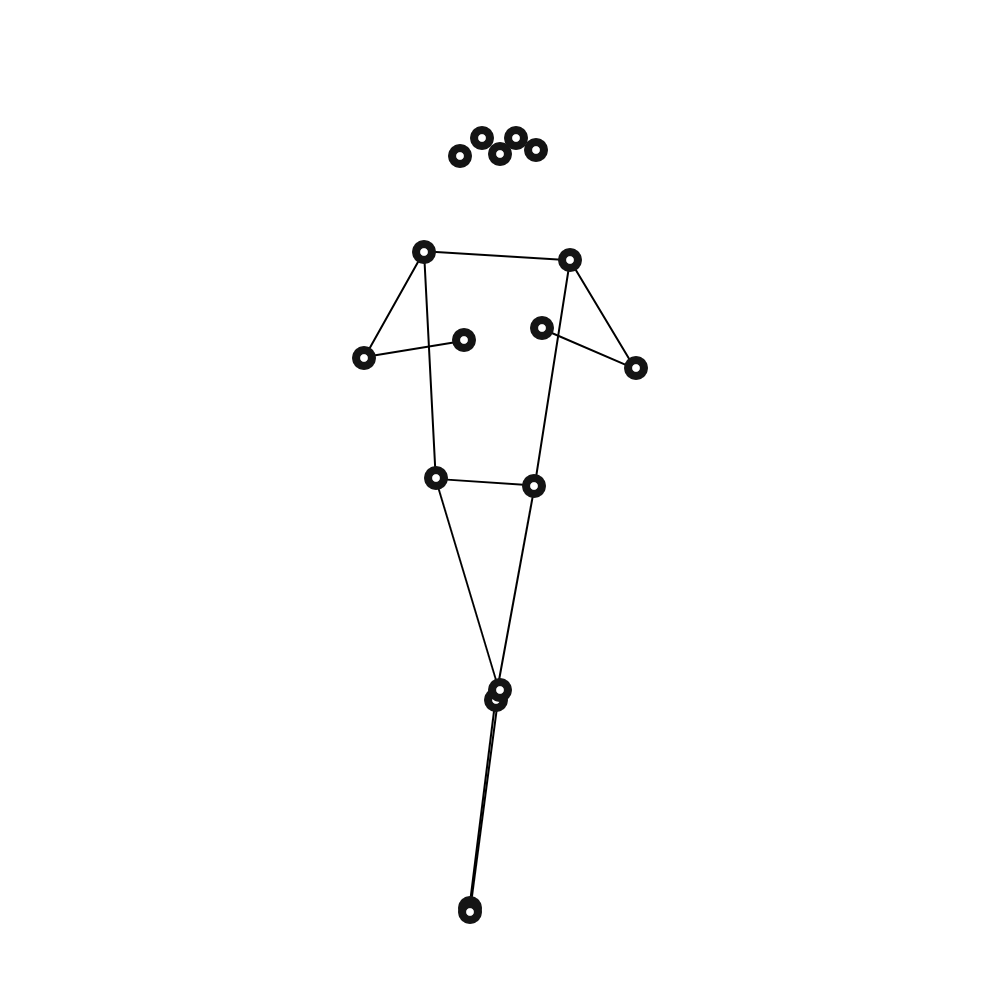}
    \includegraphics[width=0.13\linewidth,height=4cm]{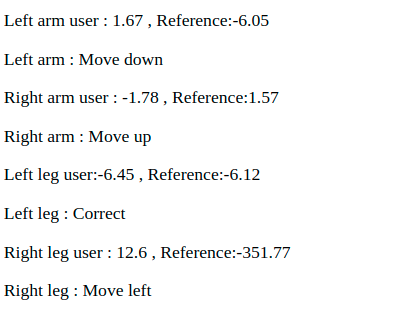}
    \hspace{0.25cm}
    \includegraphics[width=0.13\linewidth,height=4cm]{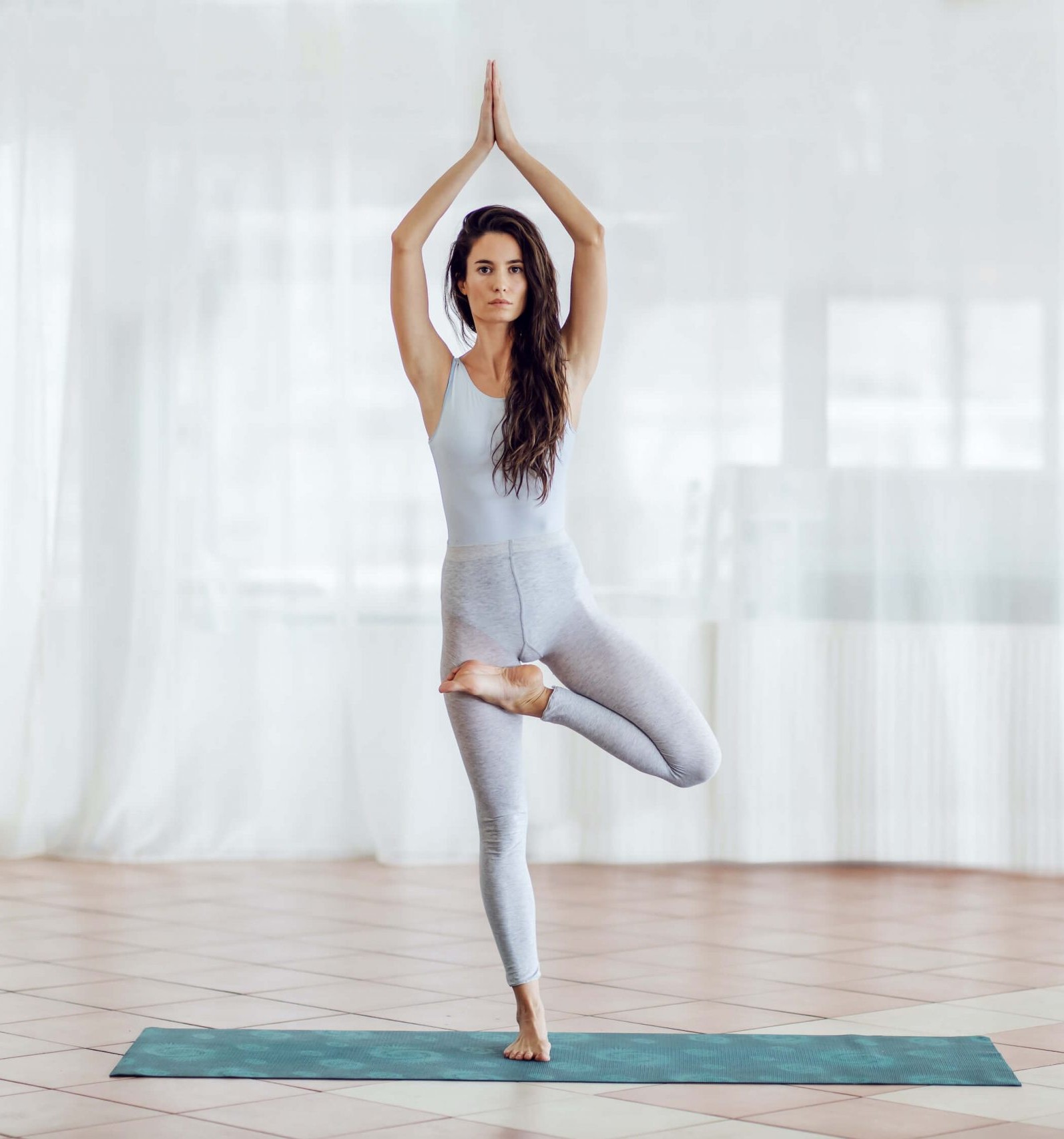}
    \includegraphics[width=0.13\linewidth,height=4cm]{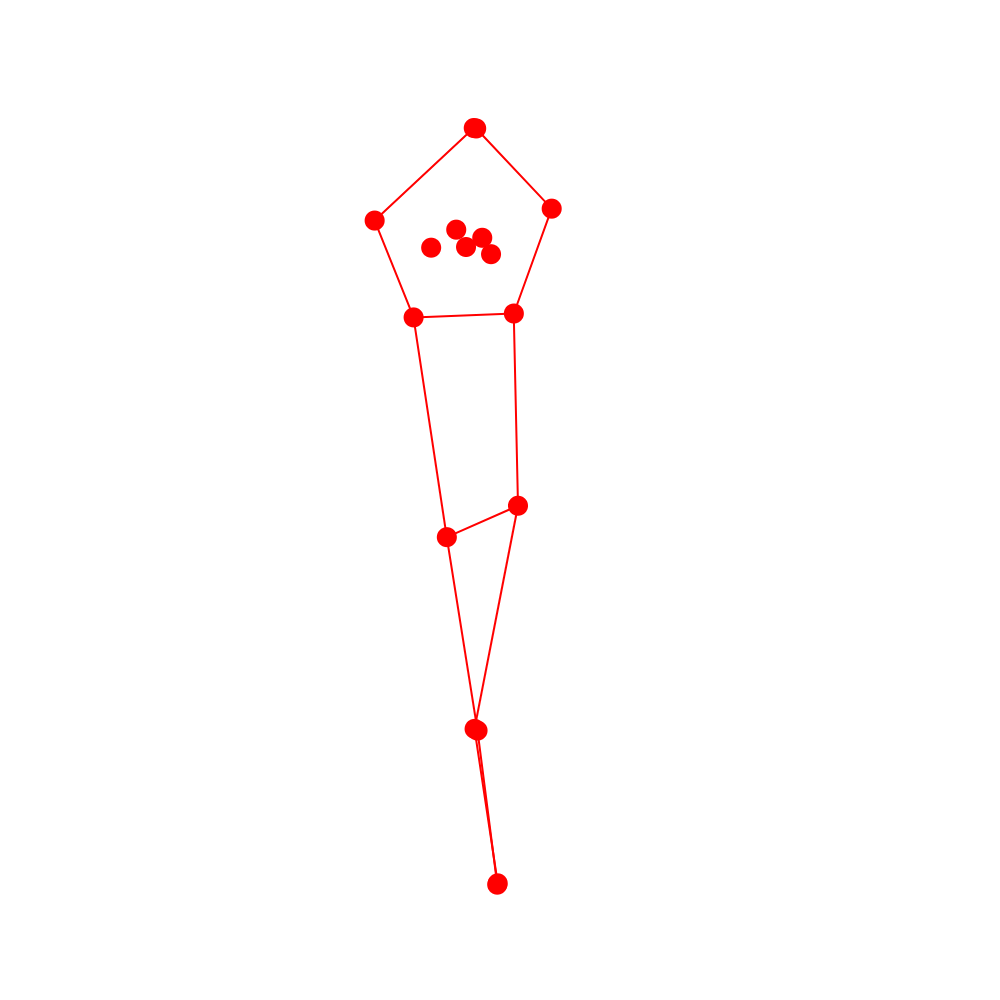}
    \includegraphics[width=0.13\linewidth,height=4cm]{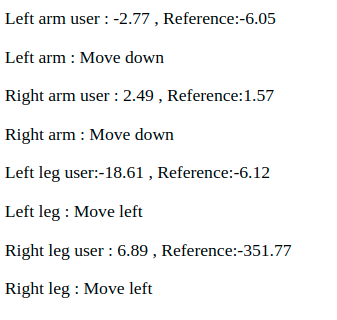} 
    \includegraphics[width=0.14\linewidth,height=4cm]{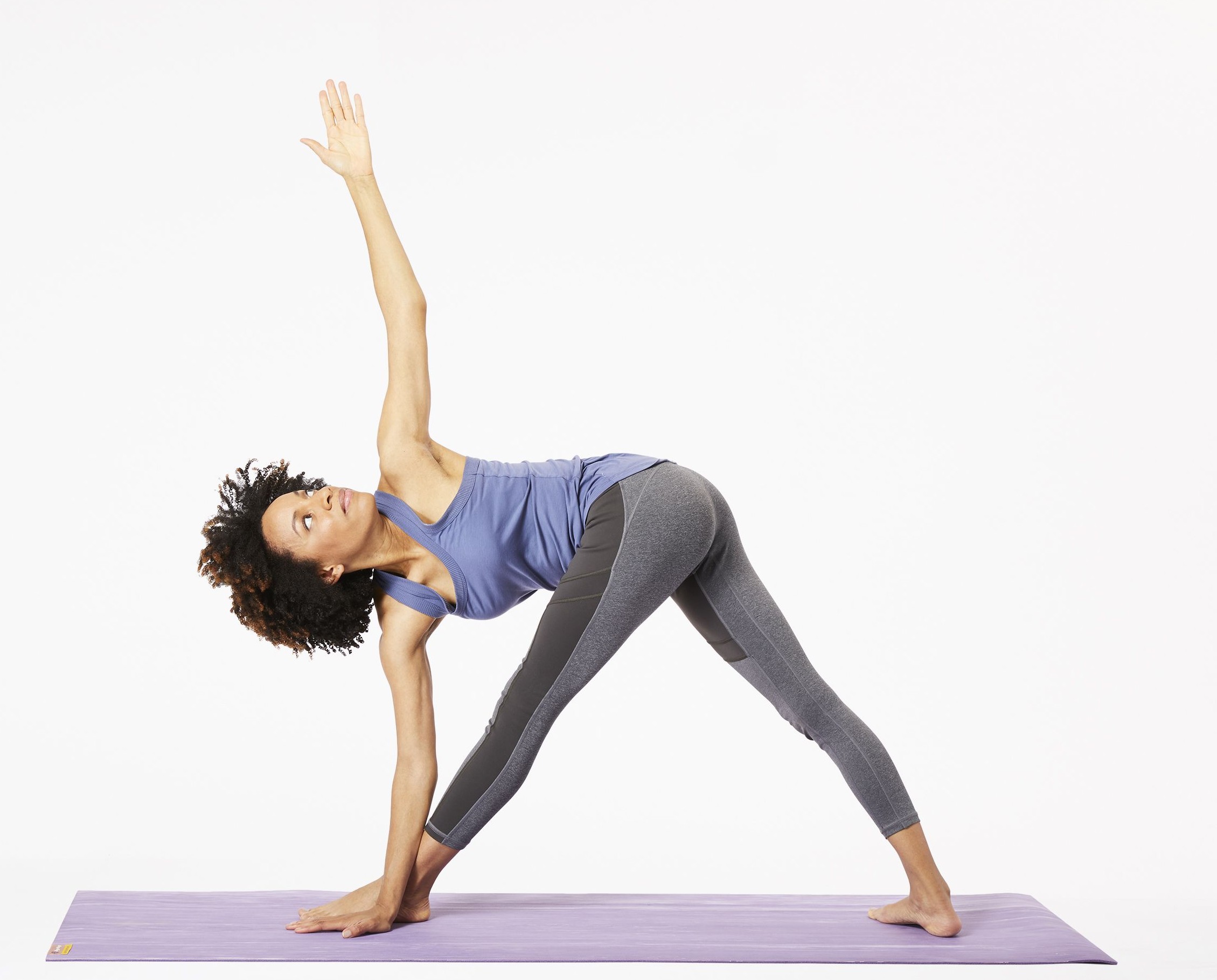}
    \hspace{0.25cm}
    \includegraphics[width=0.13\linewidth,height=4cm]{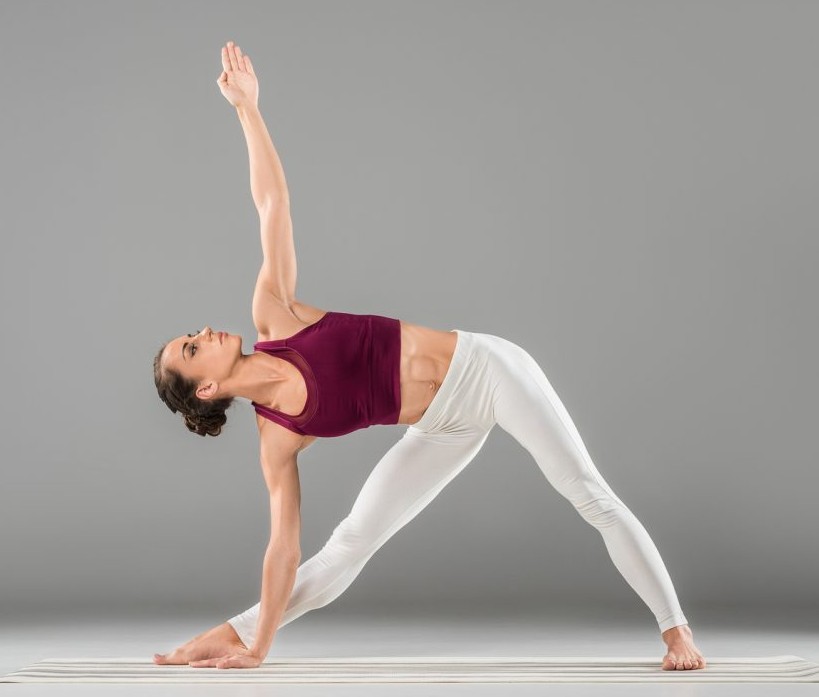}
    \includegraphics[width=0.13\linewidth,height=4cm]{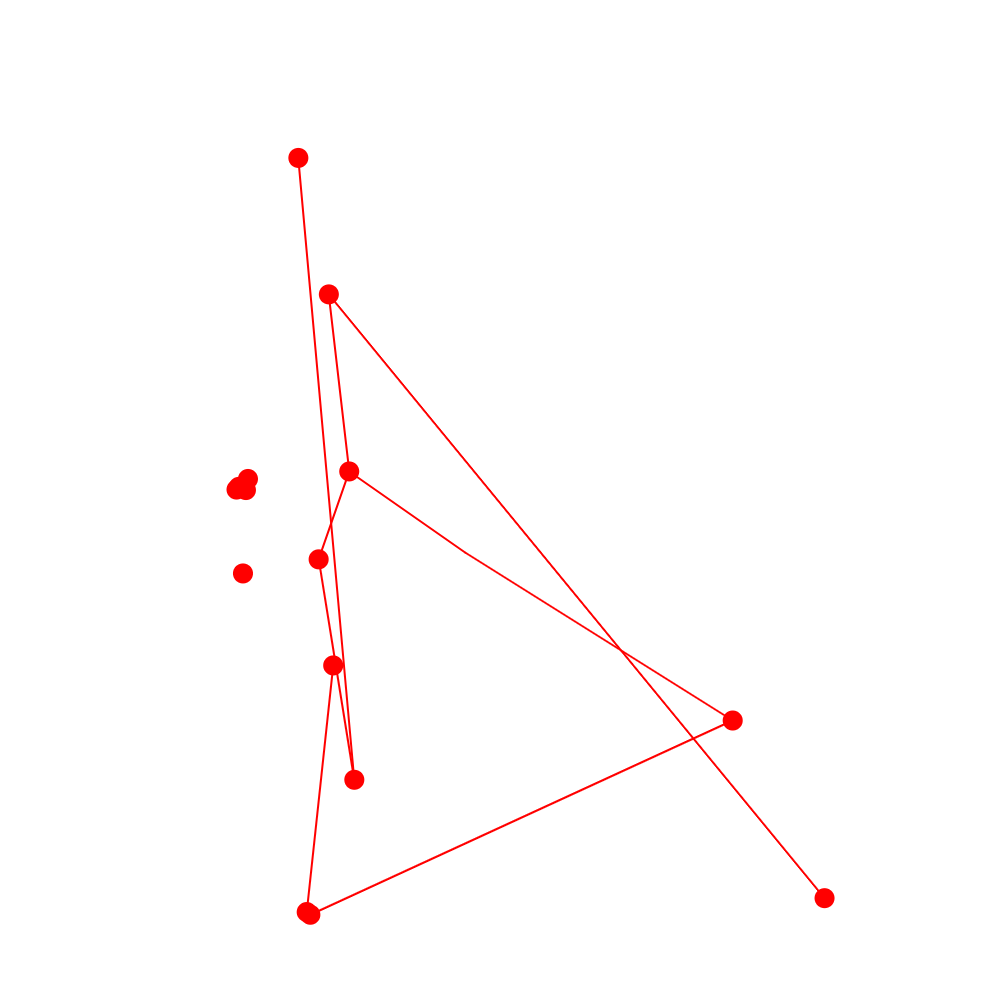}
    \includegraphics[width=0.13\linewidth,height=4cm]{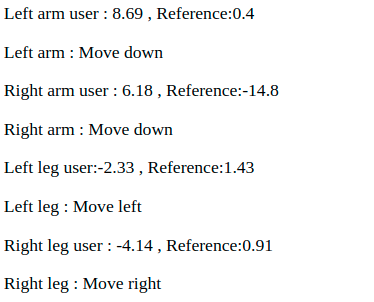}
    \hspace{0.25cm}
    \includegraphics[width=0.13\linewidth,height=4cm]{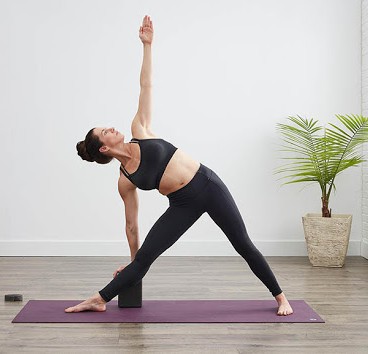}
    \includegraphics[width=0.13\linewidth,height=4cm]{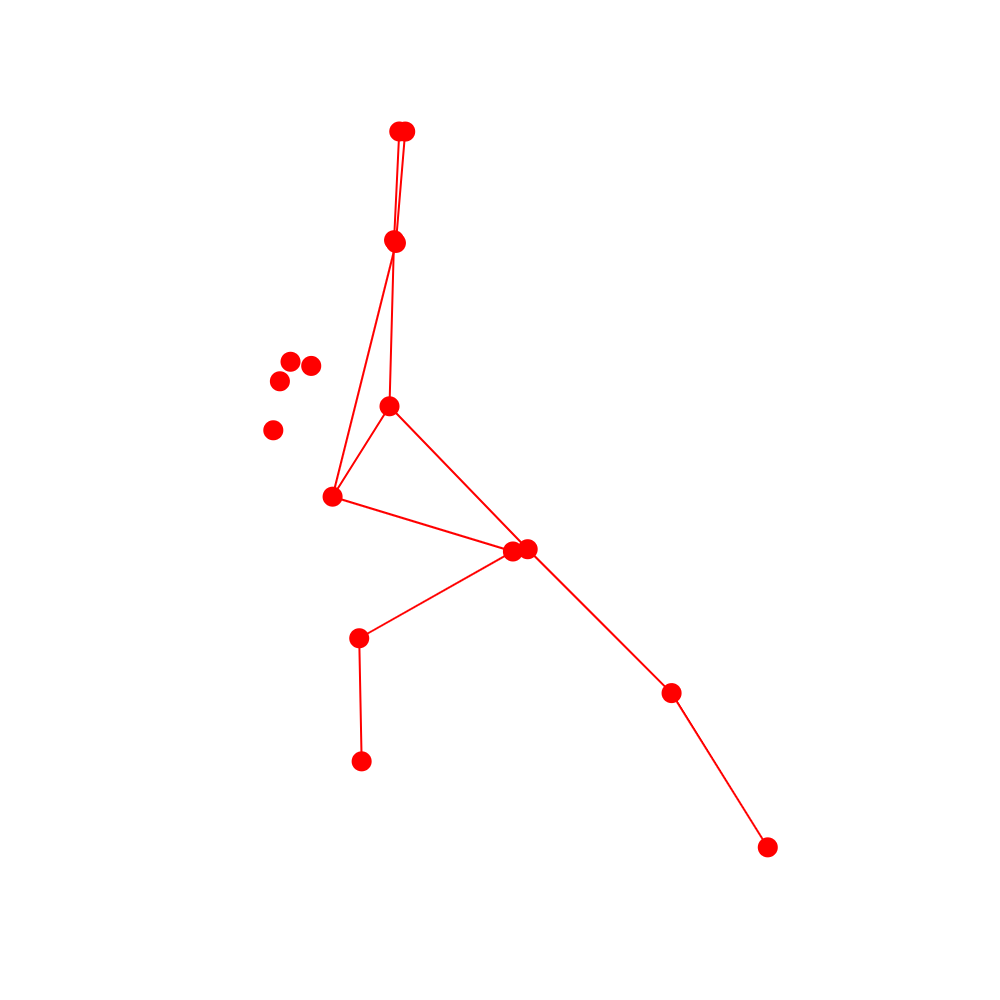}
    \includegraphics[width=0.13\linewidth,height=4cm]{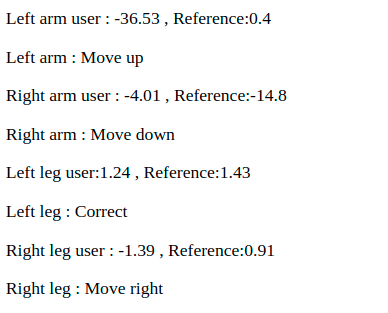}
    \caption{The results obtained from using the Fitness Tutor can be seen in this figure. The leftmost image in each row represents the referenced image. The rest of the images are the user input images, their skeleton and the suggestions obtained using pose estimation model. The slope and posture correction result is added for each user input images.}
    \label{fig:slopefig}
    
\end{figure*}

From the slope values obtained, we compare certain slopes made by body parts whether value is $\pm0.5$ correct. If the
slope made by hands, legs in certain posture is more or less correct, then the posture is right. If either of hands or legs is not in right posture, it displays a message of which part has to be corrected in order to get same posture as professional. This way there is a possibility that everyone can exercise in
right way as it is to be. This results can suggest for arms to move up/down. Similary, for both the legs the suggestions are provided to move the leg left or right according to the slope it makes. Figure \ref{fig:slopefig} represents the results obtained by training different poses.

From the results obtained, it is clear that most of the tutoring can be done in this simple way. In the figure, although the posenet generated skeleton are misaligned, the posenet generated coordinates for joints are correct. Thus we use coordinates for correcting the posture.

\section{Experimentation}
\begin{figure}[ht!]
    \centering
    \includegraphics[width=0.45\linewidth,height=5cm]{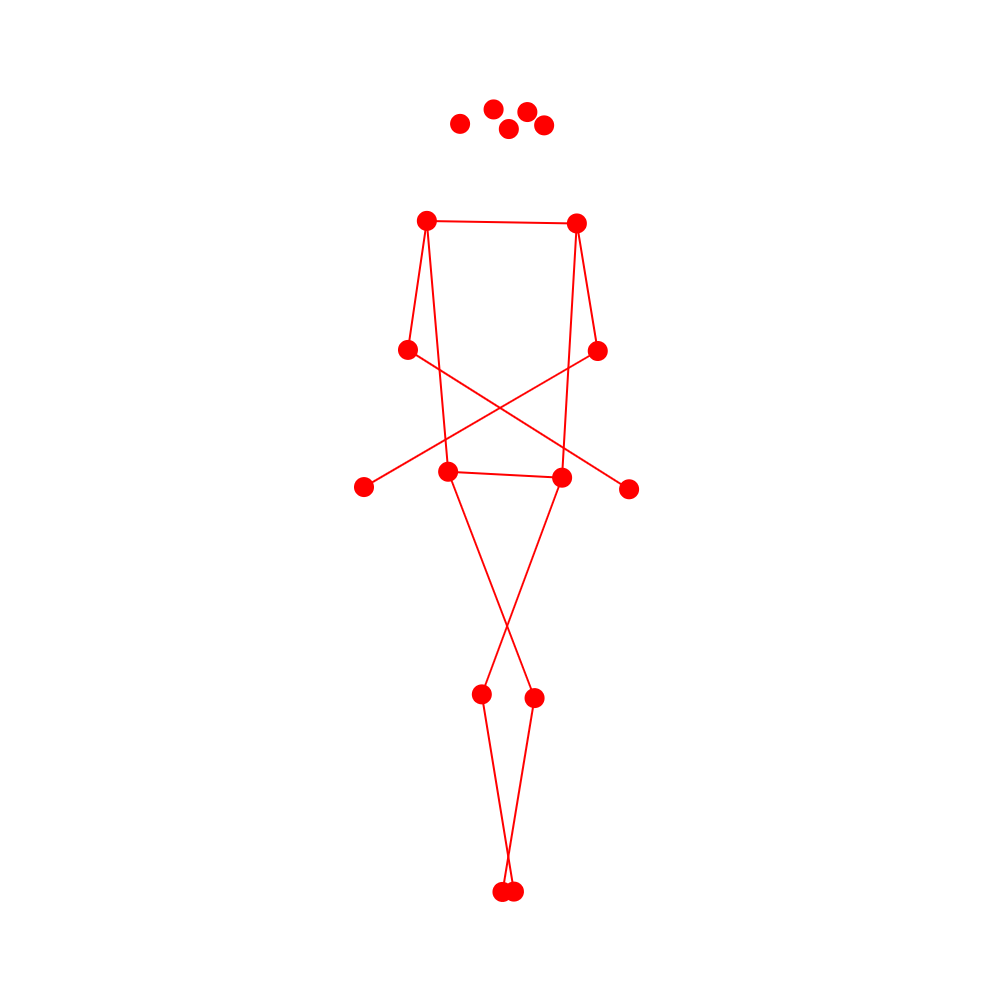}
    \includegraphics[width=0.45\linewidth,height=5cm]{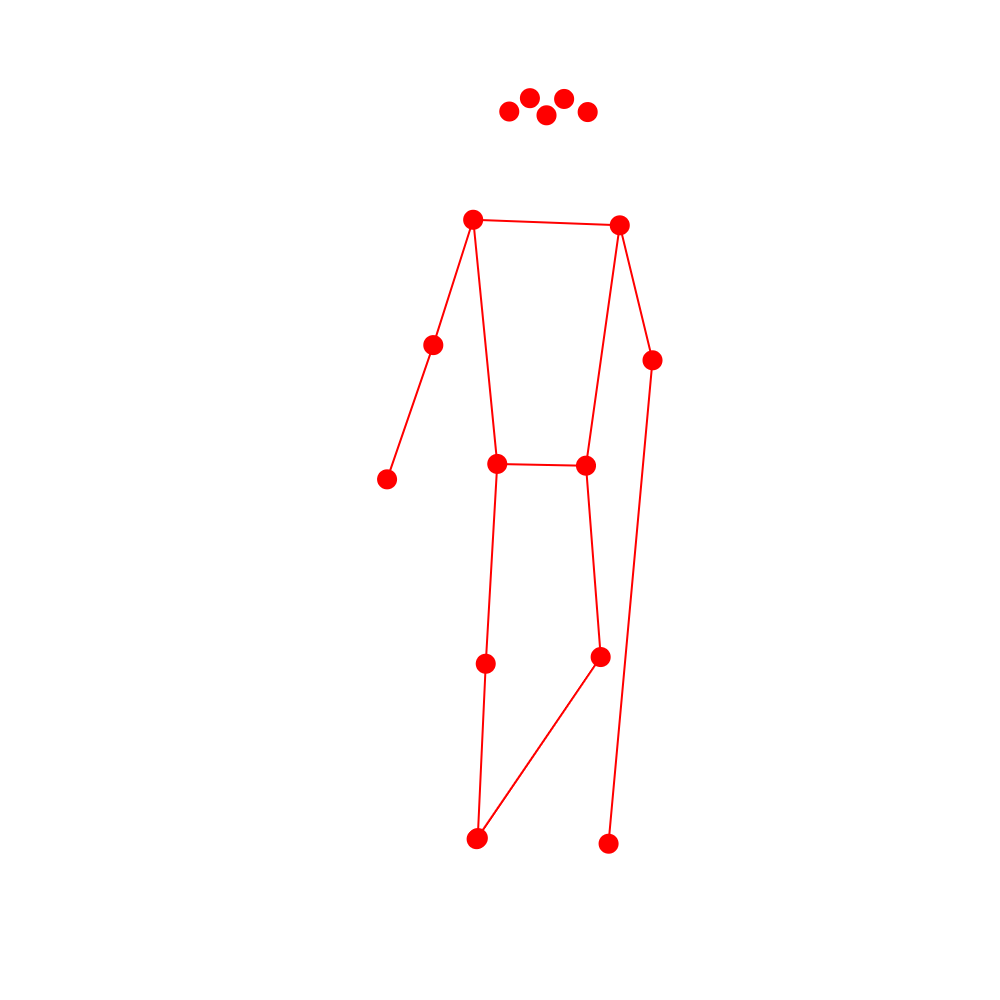}
    \includegraphics[width=0.45\linewidth,height=5cm]{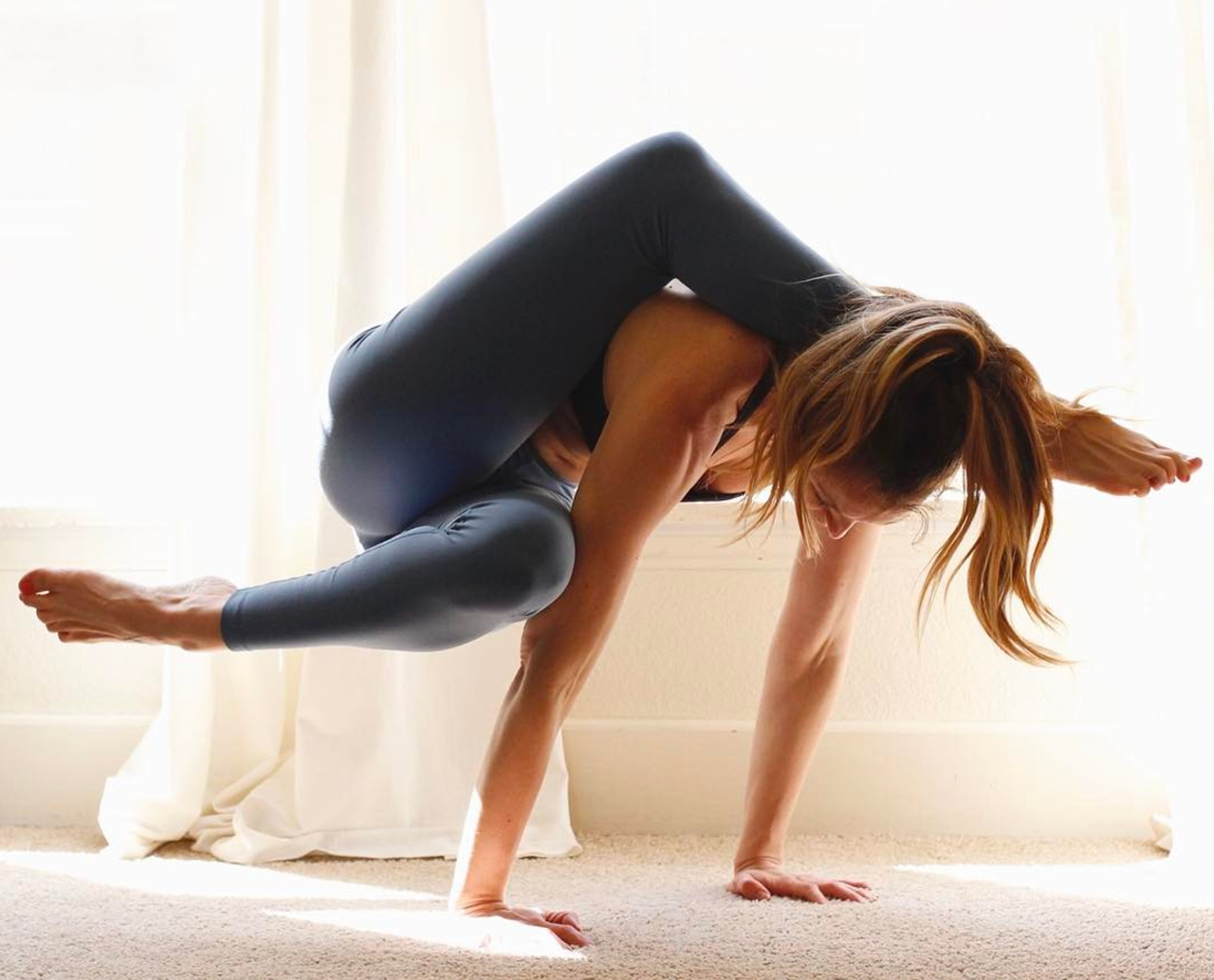}
    \includegraphics[width=0.45\linewidth,height=5cm]{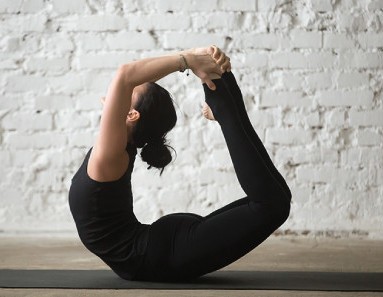}
    \includegraphics[width=0.45\linewidth,height=5cm]{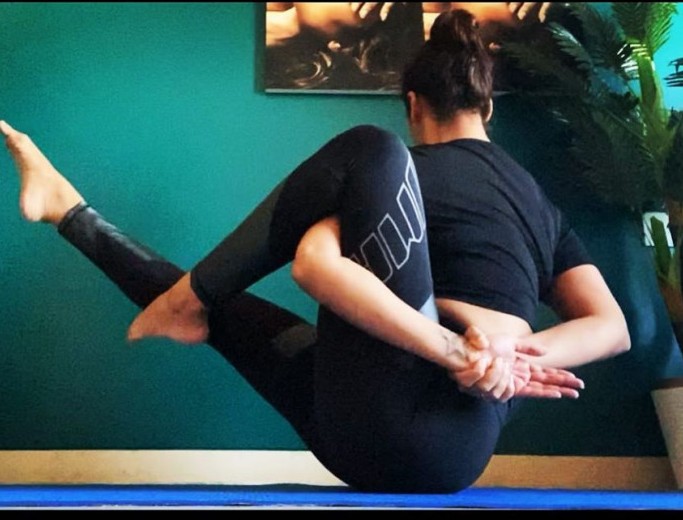}
    \includegraphics[width=0.45\linewidth,height=5cm]{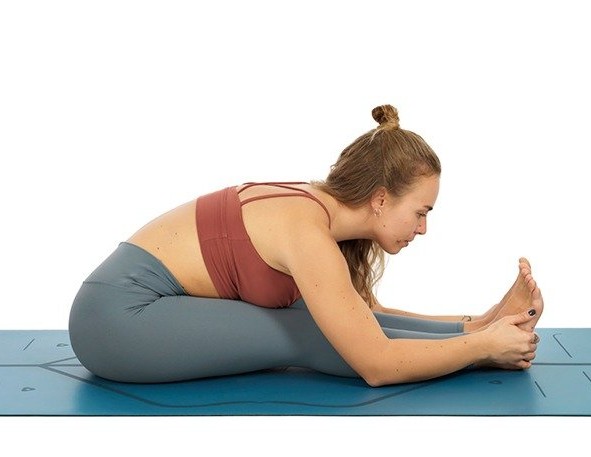}
    \includegraphics[width=0.45\linewidth,height=5cm]{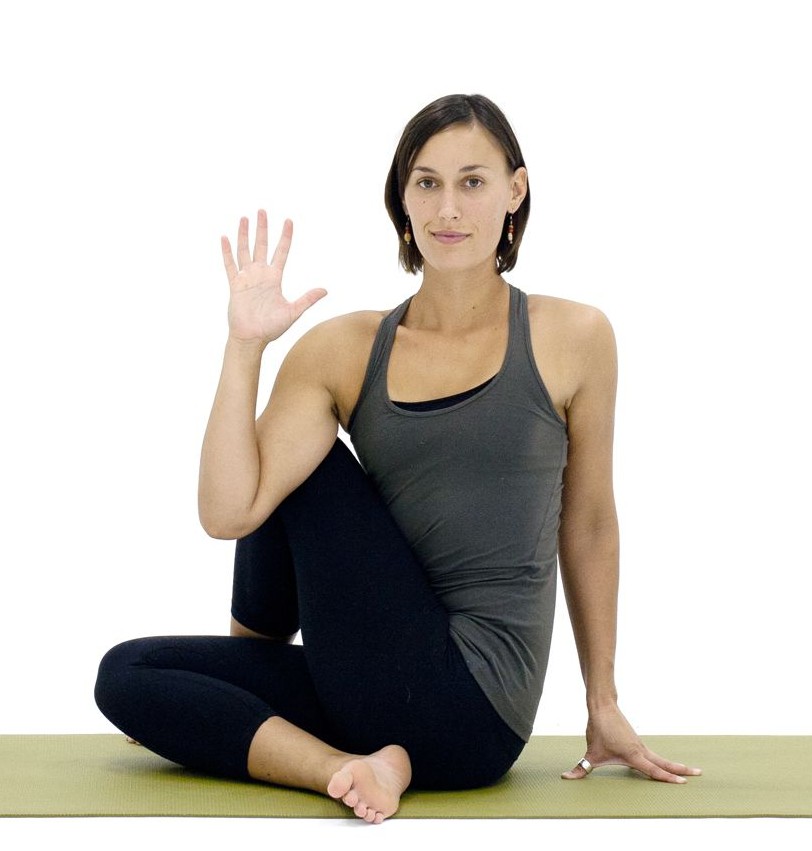}
    \includegraphics[width=0.45\linewidth,height=5cm]{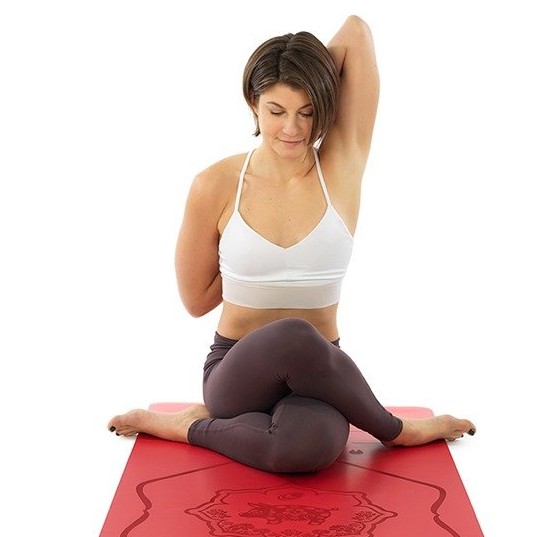}
    \caption{This picture shows the incorrect calculation of image skeleton due to wrong camera angles.}
    \label{fig:wrong}
\end{figure}

We experimented the work on images taken from Youtube videos. We compared the same exercise/asana with the other images assuming as user. The realtime suggestions provided similar to results in the table. Although most of the experimented images provide correct results, there exists some difficulties in providing a correct posture always. Some of the wrong models provided are shown in this figure \ref{fig:wrong}. Some results for referenced images or user inputs cannot be identified sue to the wrong camera angle, no proper lighting and so on. So in order to reduce the misinterpretation of images, placing the camera in right position is recommended.

Also for some exercise postures, the skeleton cannot be identified. The figure tells us that skeletons developed from these posture cannot provide a suggestion to know the posture is right or wrong. This is the drawback of implementation of pose estimation in fitness.
\section{Future Work}

There is a setback of implementing this pose estimation algorithm for certain postures shown in Figure \ref{fig:wrong}. This problem exists due to the 2D representation of body skeleton from referenced image and camera feed. This work can be extended to estimation of poses in 3D environment and mentoring in real-time.

\section*{Results}
Fitness tutor have a direct impact for human lives. This is a mentoring application that learns from 2D reference image and guides the user to mimic the posture. Slope between two vectors are used to compare the body parts of referenced image and live camera feed. In the long term our approach could enable powerful tools for better inclusion of existing 3D pose estimation models for guiding the human workouts. 



{\small
\bibliographystyle{ieee_fullname}
\bibliography{egbib}
}

\end{document}